\documentclass[journal,onecolumn]{IEEEtran}

\usepackage{hyperref}

\ifCLASSINFOpdf
\else
\fi
\usepackage{amsfonts}
\usepackage{amssymb}
\usepackage{amsmath}
\usepackage{algorithm,algpseudocode}
\usepackage{}
\usepackage{url}
\usepackage{graphicx}
\usepackage{color}
\usepackage{colortbl}
\usepackage{subcaption}
\usepackage{lineno}
\usepackage{acro}

\usepackage[
    type={CC},
    modifier={by-nc-nd},
    version={4.0},
]{doclicense}

\def\B{{\mathcal B}}
\def\C{{\mathbf C}}
\def\D{{\mathcal D}}
\def\S{{\mathcal S}}
\def\T{{\mathcal T}}
\def\R{{\mathcal R}}

\def\z{{\mathbf z}}

\newcommand{\norm}[1]{\left\lVert#1\right\rVert}

\newcommand{\argmin}{\mathop{\text{arg\,min}}}
\renewcommand{\max}{\mathop{\text{max}}}

\DeclareAcronym{cnn}{short=CNN, long=Convolutional Neural Network}
\DeclareAcronym{uav}{short=UAV, long=Unmanned Aerial Vehicle}
\DeclareAcronym{ot}{short=OT, long=Optimal Transport}
\DeclareAcronym{al}{short=AL, long=Active Learning}
\DeclareAcronym{ts}{short=TS, long=Transfer Sampling, first-long-format=\itshape}
\DeclareAcronym{svm}{short=SVM, long=Support Vector Machine}
\DeclareAcronym{iou}{short=IoU, long=Intersection-over-Union}
\DeclareAcronym{auc}{short=AUC, long=Area under the Curve}
\DeclareAcronym{mlp}{short=MLP, long=Multi-Layer Perceptron}
\DeclareAcronym{nms}{short=NMS, long=Non-Maximum Suppression}

\begin{document}
\title{Half a Percent of Labels is Enough: Efficient Animal Detection in UAV Imagery using Deep CNNs and Active Learning}
\author{Benjamin~Kellenberger~\IEEEmembership{Member,~IEEE},
		Diego~Marcos,
		Sylvain~Lobry,
		Devis~Tuia,~\IEEEmembership{Senior Member,~IEEE}
}


\maketitle
\doclicenseThis

\begin{abstract}
\textbf{Note: this is a pre-print version of work published in IEEE Transactions on Geoscience and Remote Sensing (TGRS; in press). The paper is currently in production, the DOI link will be active soon.  DOI: \href{https://doi.org/10.1109/TGRS.2019.2927393}{10.1109/TGRS.2019.2927393}.\\}
We present an \ac{al} strategy for re-using a deep \ac{cnn}-based object detector on a new dataset. This is of particular interest for wildlife conservation: given a set of images acquired with an \ac{uav} and manually labeled gound truth, our goal is to train an animal detector that can be re-used for repeated acquisitions, e.g. in follow-up years. Domain shifts between datasets typically prevent such a direct model application. We thus propose to bridge this gap using \ac{al} and introduce a new criterion called \ac{ts}. \ac{ts} uses Optimal Transport to find corresponding regions between the source and the target datasets in the space of \ac{cnn} activations. The \ac{cnn} scores in the source dataset are used to rank the samples according to their likelihood of being animals, and this ranking is transferred to the target dataset. Unlike conventional \ac{al} criteria that exploit model uncertainty, \ac{ts} focuses on very confident samples, thus allowing a quick retrieval of true positives in the target dataset, where positives are typically extremely rare and difficult to find by visual inspection. We extend \ac{ts} with a new window cropping strategy that further accelerates sample retrieval. Our experiments show that with both strategies combined, less than half a percent of oracle-provided labels are enough to find almost 80\% of the animals in challenging sets of \ac{uav} images, beating all baselines by a margin.
\end{abstract}


\markboth{PREPRINT, full version: \href{https://doi.org/10.1109/TGRS.2019.2927393}{10.1109/TGRS.2019.2927393}}{Kellenberger et al.: \mytitle}

\IEEEpeerreviewmaketitle

\section{Introduction}

\IEEEPARstart{R}{epeated} wildlife censuses provide an invaluable tool for ecologists to count animals, monitor population health and stem threats from poaching incidents~\cite{Hodgson2013,Yang2014}. Population densities and spatial locations of big mammals are constantly fluctuating, and having up-to-date information on where and how many individuals are found may be decisive for grazing needs estimation, or for the success of anti-poaching means. Hence, authorities of national parks and game reserves require animal census tools that are fast, reliable, and suitable for repeated applications over time.

Traditional censuses using manual surveys from manned helicopters~\cite{bayliss1989distribution,norton1978counting} are steadily replaced by approaches using \acp{uav}~\cite{Linchant2015}. \acp{uav} are inexpensive, remotely-controlled aircrafts that can be equipped with small payloads like compact imaging cameras. Latest studies have shown censuses based on \ac{uav} imagery to yield superior accuracy compared to human surveys~\cite{Hodgson2018}. They are especially appealing when combined with methods from machine learning and computer vision~\cite{Rey2017,Kellenberger18}, in particular with object detectors~\cite{Ren2015,Redmon2016,Redmon2017} employing deep \acp{cnn}~\cite{Krizhevsky2012,LeCun2015}: such models allow fast scans of the thousands of images \acp{uav} produce over game reserves of average sizes (i.e., hundreds of square kilometers), thereby alleviating the tedious work of manual photo-interpretation. This is particularly important in real-world scenarios where animals are a rare sight and images are dominated by empty background.

\begin{figure}
\centering
\includegraphics[width=0.75\linewidth]{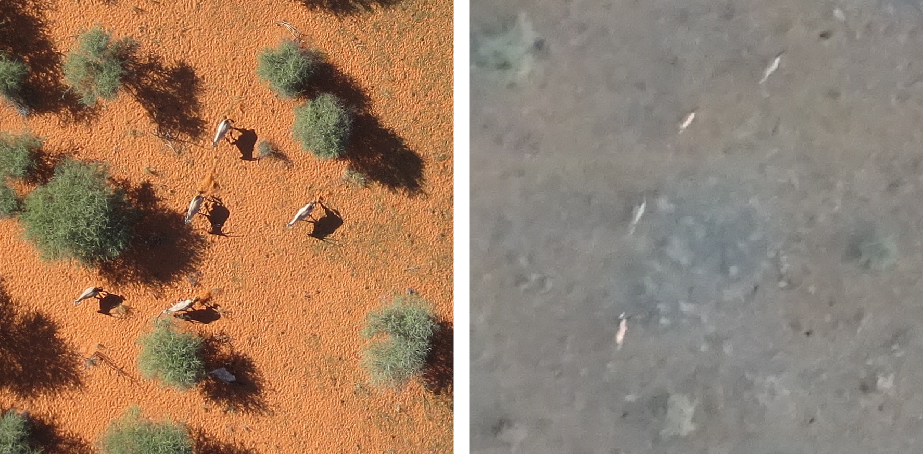}
\caption{Examples from the Kuzikus dataset (see Section~\ref{sec:dataset}) from 2014 (left) and 2015 (right). It is trivial for humans to identify the animals in either image, but a model trained on only one dataset is likely to fail when predicting animals in the other.}
\label{fig:domainShift}
\end{figure}

However, these models are typically trained on a single dataset and quickly break down in accuracy when applied to others. This problem is known as domain shift and denotes the inherent differences present between acquisitions~\cite{tuia2016domain}. For example, the image crops in Figure~\ref{fig:domainShift} are from the same game reserve, but are clearly very different in characteristics. For a human, it is trivial to locate the animals in either scene; a machine trained on only one set, however, is likely to fail when run on the other. In practice, even if it still finds most of the animals (high recall), such a model is likely to also produce false alarms everywhere in the background (low precision).

In the literature, this discrepancy is commonly solved by means of domain adaptation~\cite{tuia2016domain}, where a model trained on one dataset (\emph{source} domain) is modified to also work on another (\emph{target} domain). Multiple approaches have been proposed to this end, including unsupervised ones that only consider the images of the target domain, and semi-supervised methods that further assume the presence of a small number of labels (animal positions) in the target domain.

As soon as the dominance and appearance variability of the background class get very high, a certain degree of supervision becomes unavoidable. This raises the question on how the few target labels can be obtained that are required to this end. A naive approach could require human operators to sift through hundreds of images before encountering an animal, which is highly inefficient and can lead to fatigue. This in turn likely causes erroneous labels, and hence missed targets and loss of accuracy. To this end, multiple studies have resorted to \acreset{al}\ac{al}~\cite{Tuia2011}. In \ac{al}, a machine (model) works hand-in-hand with a so-called oracle (typically a human expert) and exploits their knowledge by issuing queries for ground truth whenever it encounters a particularly relevant data sample.

The notion of relevance conventionally refers to the usefulness of a sample to the final model performance on the target dataset~\cite{Tuia2011}. Multiple \ac{al} criteria have been proposed~\cite{settles2012active}: for example, uncertainty sampling methods like Breaking Ties~\cite{luo2005active} exploit the model's confidence on samples; model-specific approaches like margin sampling for \acp{svm}~\cite{schohn2000less}, expected model change~\cite{cai2013maximizing}, or the recent Bayesian \acp{cnn}~\cite{gal2017deep} make use of individual model properties to establish a sample ranking. They all seek for a prioritization of samples that lead to the highest performance of the underlying model with a small, given number of queries to the oracle.

In the case of animal censuses, however, things are different: instead of improved model generalization capability, park rangers are primarily interested in \emph{locating the animals} in the new dataset. In this context, established criteria are likely to break down, since they tend to sample in areas where the detector is uncertain and the likelihood of obtaining a true positive is very low. Finding animals thus requires an \ac{al} criterion that works in the opposite direction by prioritizing predictions that are most certainly true positives (instead of low-confidence samples).

This means that two deviating objectives need to be met: fast animal localization on the one hand, but also some model improvement on the other. The latter refers to the interactive part of the \ac{al} adaptation paradigm: one could just train a detector on source, apply it once on the target images and then use a criterion in one go. This is known as ``one-shot'' \ac{al}~\cite{santoro2016one}. However, we argue that using the newly obtained labels at every \ac{al} iteration to update the model and re-predict candidates can lead to increasingly higher quality predictions, and thus to higher chances of true positives retrieval.

In this paper, we therefore present a novel strategy that allows finding as many animals as possible with minimal labeling effort in a new \ac{uav} acquisition, using an available source dataset and detector, \ac{al} and an oracle in the loop. In detail, the contributions of this work are as follows:
\begin{enumerate}
\item We introduce an \ac{al} criterion that, unlike conventional approaches, seeks to maximize the encounter probability of (rare) true positive candidates in the target domain.
\item Furthermore, we present a window cropping strategy that allows obtaining more labels per query while also being more intuitive for human annotators to label.
\item We provide an evaluation, comparison and ablation study on a UAV dataset of two distinct acquisitions characterized by domain shifts. Results show that, when using the proposed smart sampling strategy, it is possible to retrieve 80\% of the animals by screening only half a percent of the acquired dataset.
\end{enumerate}

The rest of this paper is organized as follows:

Section~\ref{sec:proposedMethod} explains the main procedure, including the \ac{al} criterion denoted as ``Transfer Sampling'' (Section~\ref{sec:TransferSampling}), as well as the window cropping strategy (Section~\ref{sec:windowCropping}). We put the model to the test in Section~\ref{sec:experiments}, results of which we show and discuss in Section~\ref{sec:results}. Finally, we draw conclusions from our work in Section~\ref{sec:conclusion}.
\newpage

\section{Proposed Method} \label{sec:proposedMethod}

\begin{figure*}[!ht]
\centering
\includegraphics[width=0.75\linewidth]{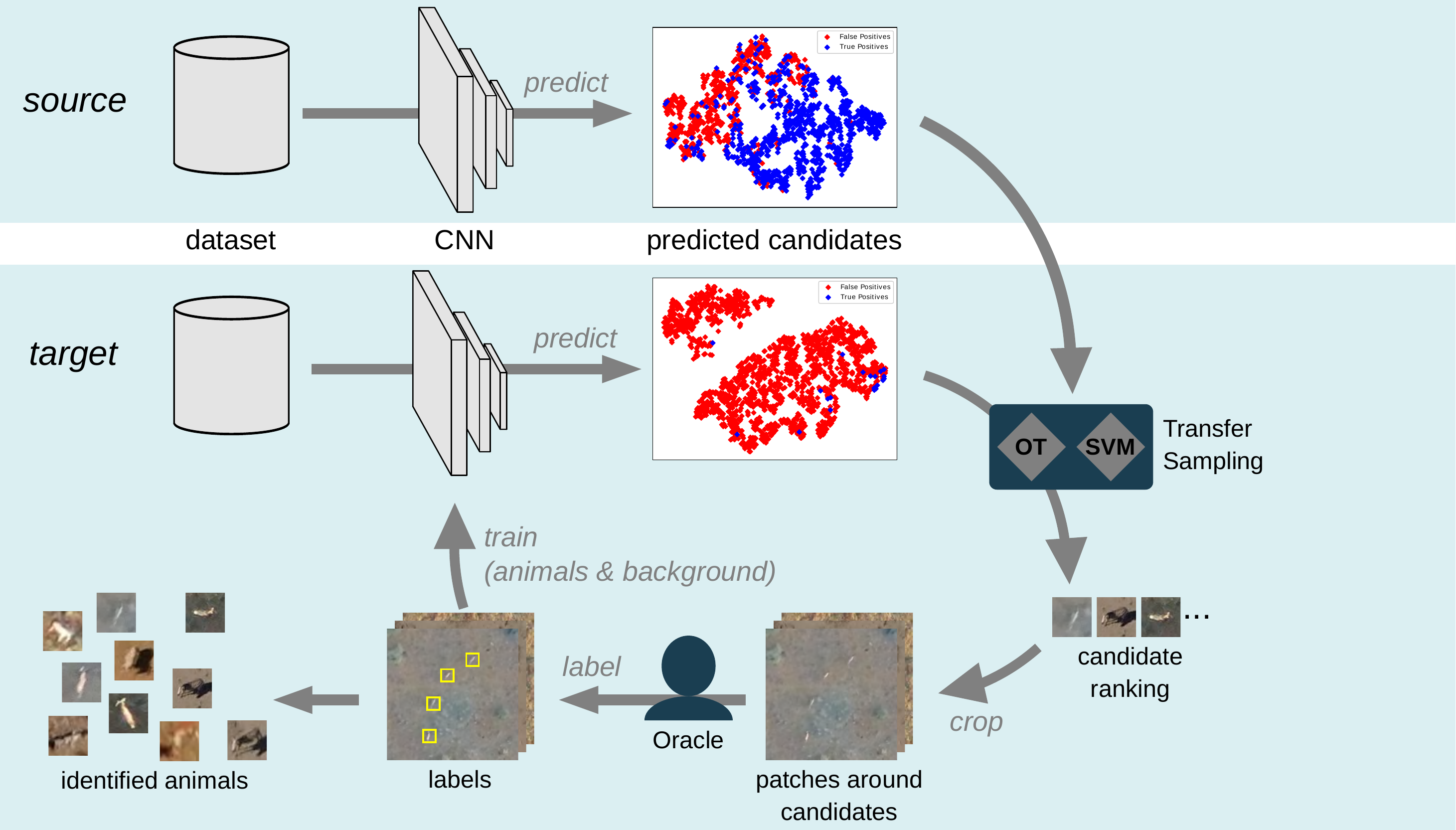}
\caption{Overview of the proposed workflow. We first predict candidates in the source dataset using the original, unadapted \ac{cnn} (top row). We do the same on the target dataset using the current \ac{cnn} (below). These serve as inputs for our \ac{ts} strategy (right), which ranks the source samples with an \ac{svm} and transports the ranking to the target candidates via \ac{ot}. These ranked target candidates serve as anchor points for patches, which in turn form the query data to the oracle (bottom). The latter provides labels for subsequent \ac{cnn} training (completion of the \ac{al} loop).}
\label{fig:overview}
\end{figure*}

Figure~\ref{fig:overview} provides an overview of the proposed interactive domain adaptation workflow. As a precondition, it assumes the presence of a source dataset and an object detector (a deep \ac{cnn} in our case) that has been trained on it. The model and its parameters are initially copied to extract features at every location in the images from the target domain. The distributions of these features in the source and target domain are then matched using \ac{ot}~\cite{cuturi2013sinkhorn}, which allows transferring the source ground truth labels to the target domain. This provides a means of confidence prediction for the target samples, which can then be verified by an expert oracle.

By iteratively querying the oracle to provide ground truth labels for the most likely true positives, the model can then gradually be fine-tuned to the target domain and provides better proposals in the following \ac{al} iterations, while making sure that we minimize the tedium of the prospective oracle. We limit experiments to ground truth-based (simulated) oracles in this work, but present a further strategy to accelerate and facilitate manual annotations in upcoming extensions with human annotators, which we denote as ``window cropping''. In the rest of this section, we discuss the two key components of the adaptation process: the proposed \ac{al} criterion for sample selection/ranking (Section~\ref{sec:TransferSampling}), and the window cropping strategy (Section~\ref{sec:windowCropping}).

\subsection{Transfer Sampling} \label{sec:TransferSampling}

As outlined above, our main interest lies in quickly locating animals in the target domain. Starting from the set of locations where the source model predicts more than 10\%\footnote{With 10\% confidence we typically obtain recalls of 90\% without having an excessive number of false positives.} chance of animal presence (denoted as \emph{candidates} hereafter), we want to find those that are most likely to be true positives with the proposed \ac{al} criterion ``Transfer Sampling'' (\ac{ts}).

In \ac{ts}, we leverage the model's (higher) performance in the source domain and transfer this knowledge to the target samples. This is based on the assumption that the ``best'' predictions in source (i.e., the true animals) are clustered together in the feature space of the last layer of the \ac{cnn}, and that an equivalent region can be found in the target domain that is similarly relevant.

\begin{figure}
\centering
\includegraphics[width=0.65\linewidth]{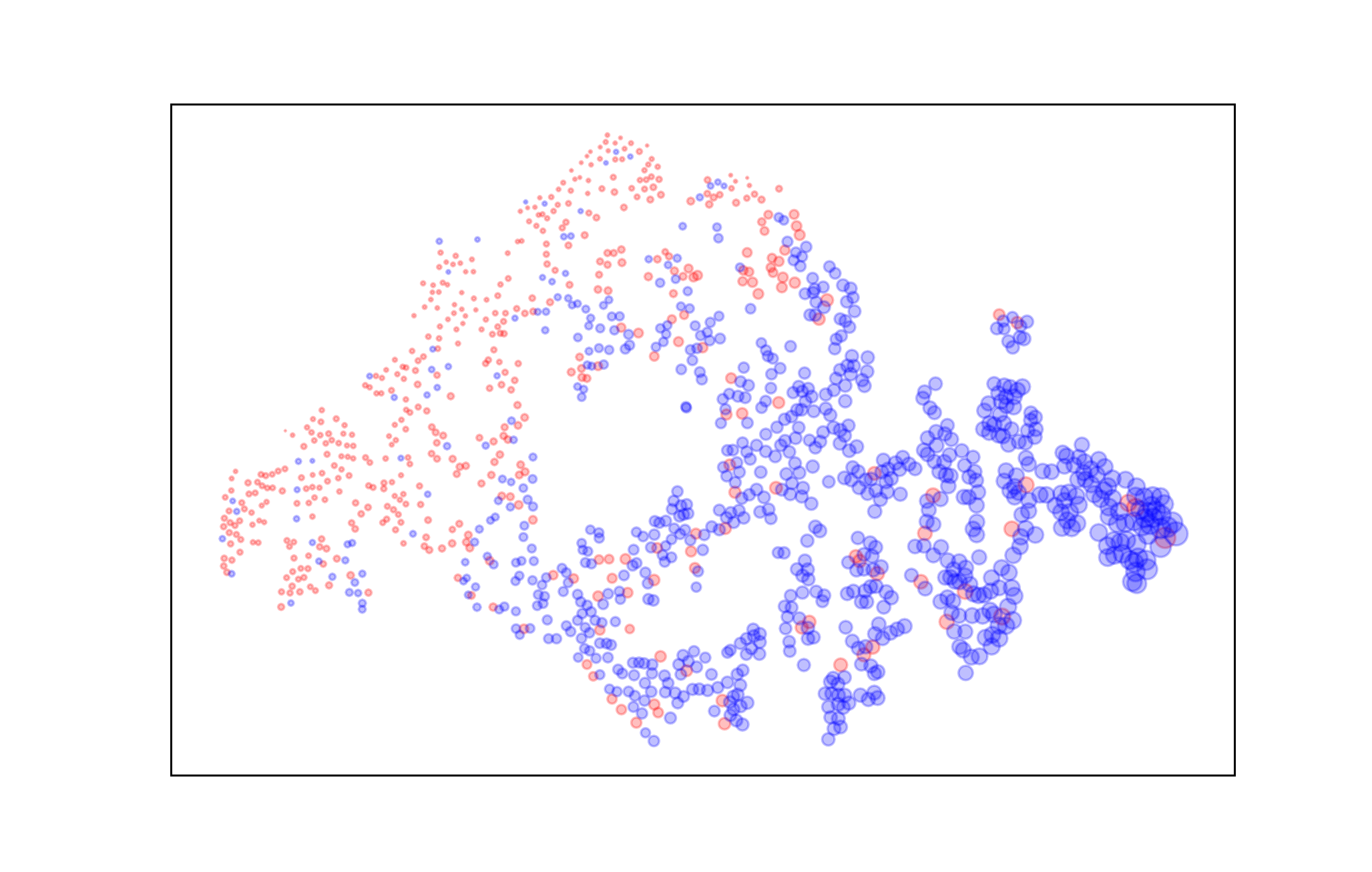}
\caption{Source dataset candidates with animal confidence of 0.1 or more, projected using t-SNE~\cite{VanDerMaaten2008}. Blue samples were predicted correctly (true positives), red samples denote false alarms. The marker size indicates the distance to the \ac{svm} hyperplane (larger = further into the true positives region).}
\label{fig:extremeSampling_graphical}
\end{figure}

The challenge in this context is the imbalance between animals and background, combined with a likely excessive number of false detections made by the source model in the target dataset. To find the animals quickly and keep the annotators' motivation high, we thus need to prioritize target candidates whose corresponding source predictions were indeed true positives. We therefore consider sampling according to similarities in the \ac{cnn}'s feature space, spanned by the deep animal detector in the source and target domains. Figure~\ref{fig:extremeSampling_graphical} shows all samples in the source domain that were predicted by the source detector as ``animals''. Although the model still makes a number of false positives (red), a good majority of true positives (blue) is consistently clustered in one region of the feature space. In such a scenario, it therefore makes sense to start sampling in the far right blue area, since these are the furthest away from the transition to the false positives. We thus have two tasks to solve: \textit{(i.)} numerically identify regions in the source domain feature space that most likely contain true animals, and \textit{(ii.)} locate the same corresponding regions in the target domain.

For the first task, we resort to a margin-based auxiliary classifier to get a surrogate measure of sample certainty. In detail, we train an \ac{svm}~\cite{cortes1995support} on the full set of source candidates and then use it to rank the candidates by their distance to the separating hyperplane. This gives us an order that prioritizes samples as far away from the decision boundary as possible (the hyperplane distance is given as the marker size in Figure~\ref{fig:extremeSampling_graphical}), which in turn makes sure that the most trustworthy candidates per source domain are sampled first. This strategy is conceptually close to margin sampling commonly used in \ac{al}~\cite{schohn2000less}, but we use it to focus the sampling on the \emph{most} certain areas of the positive class, rather than the least certain ones.

The second task then consists of transferring the ranks to ``similar'' target samples. The intuition here is that both source and target candidates follow similar, mappable distributions, and we therefore need to find a way to establish an explicit source-to-target correspondence: given a predicted animal in the source domain, we want to know the predictions in the target domain that match to it. However, due to domain shifts a simple nearest neighbor search is likely to induce noise.

We propose to instead find this mapping using \acreset{ot}\ac{ot}. \ac{ot} finds a correspondence between two distributions that is optimal with respect to a global cost~\cite{Courty2015}. It does so by calculating and minimizing for the Wasserstein distance, also known as the earth mover's distance. This distance quantifies the difference between the two distributions as a product of their data similarities and individual distances. The intuition behind this idea is that parts of the two distributions might be similar by some measure, but far apart with respect to their ``location'' within the distributions. In the case of discrete distributions like ours ({\em i.e.}, the distributions are constituted by individual predicted animal candidates), this means that two candidates from each distribution (resp. domain) only get associated with each other if they are similar by some measure \emph{and} lie in similar areas of their respective distributions. In the following, we therefore assume the source and target domains to be represented by the discrete probability distributions $\mu_{\S}$ and $\mu_{\T}$:
\begin{equation}
\mu_{\D} = \sum_{i=1}^{n_\D}{p_i^{\D}\delta_{\z_i^\D}} \text{ for $\D \in \{\S,\T\}$},
\label{eq:marginal}
\end{equation}
Here, the sum over all $n$ locations (predicted candidates) of the domain $\D$, either source ($\S$) or target ($\T$), defines the discrete distribution. $\delta_i^{\D}$ denotes the Dirac at location $\z_i^\D \in \mathbb{R}^d$, with $\z_i^\D$ being the $i$th candidate's $d$-dimensional feature vector as predicted by the \ac{cnn}. $p_i^{\D}$ is the empirical probability per sample, to which we always assign the value $p_i^{\D} = 1/n_{\D}$.

This allows us to define the \ac{ot} objective for the two discrete source and target distributions: to find a set of explicit links between all the individual source and target locations that match well. To this end, \ac{ot} creates a sparse matrix $\gamma$ of size $n_{\S} \times n_{\T}$, where $n_{\S}$ (resp. $n_{\T}$) is the number of samples in the source (resp. target) domain. $\gamma$ contains non-zero values wherever specific source and target locations ``match''. This match is defined as the link contributing to a global cost $\C$ for the two samples being minimal. Intuitively, establishing a link between a source and a target sample that both lie in similar regions in the feature space induces a lower cost than if they were e.g. in opposite regions. Numerically, the optimal solution to that, i.e. the optimal \emph{transport plan}, can be obtained as follows:
\begin{equation}
\label{eq:ot}
\gamma^* = OT(\mu_\S,\mu_\T) = \argmin\limits_{\gamma \in \B}\langle \gamma, \C \rangle_F,
\end{equation}
where $\langle \cdot,\cdot \rangle_F$ is the Frobenius dot product and $\C$ the cost matrix of size $n_{\S} \times n_{\T}$. $\C_{ij}$ is the cost to move a unit amount from $z_i^\S$ to $z_j^\T$ (at source and target locations $i$ and $j$, respectively). $\B$ is the so-called transportation polytope, {\em i.e.} the set of all possible, positive matrices with prescribed marginals $\mu_\S$ and $\mu_\T$. In other words, $\B$ comprises all combinations of transport links between all $n_{\S}$ source and $n_{\T}$ target samples. For the cost term $\C$, a commonly used choice is the $\ell_2$ norm between samples~\cite{Courty2015}. We follow this approach, as we found it to work well in our setting involving \ac{cnn} features.
Equation~\eqref{eq:ot} can be formulated as a linear program, and further be solved efficiently with simplex-based algorithms as well as group regularizations~\cite{cuturi2013sinkhorn,Courty2015}. This then gives us the optimal transport plan $\gamma^* \in \mathbb{R}^{n_{\S} \times n_{\T}}$, which provides explicit correspondences between individual source and target samples that are sound with respect to the whole distributions. We note that in general, and specifically when $n_{\T}>n_{\S}$, the coupling may yield one-to-many linkages, and many-to-one in the inverse case, but is always sparse thanks to the constraint that the source and target marginal probabilities (Equation~\eqref{eq:marginal}) must sum to one.

\begin{figure}
\centering
\includegraphics[width=0.75\linewidth]{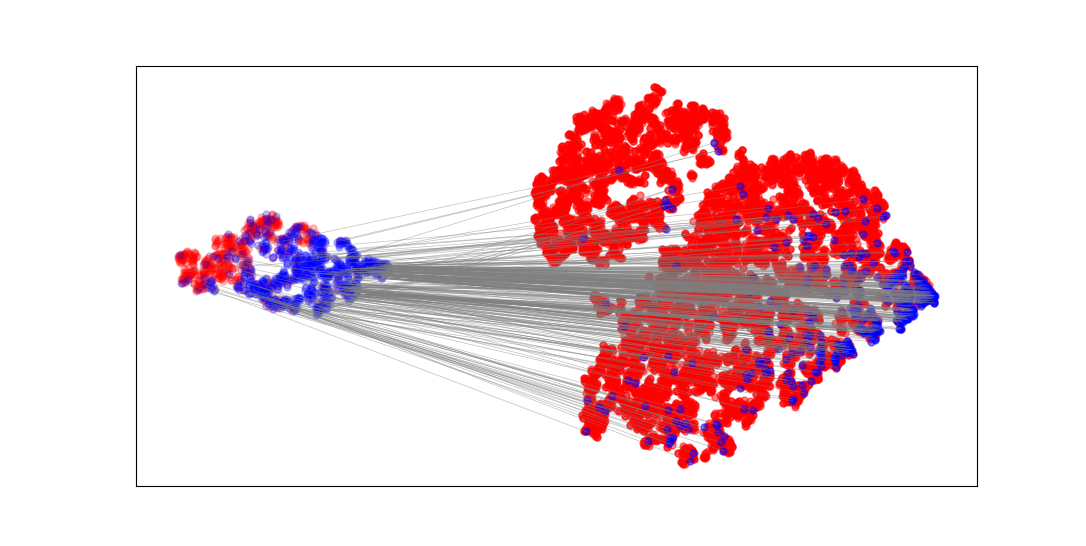}
\caption{A subset of predicted locations in the source (left) and target (right) domain training sets. Blue samples were predicted correctly (true positives) and red samples are false positives. Gray lines denote the correspondences found by \ac{ot} for all the correctly predicted target samples. Note that, despite the imbalance and higher number of false alarms in the target set, the \ac{ot} correspondences are globally consistent.}
\label{fig:OTmapping}
\end{figure}

We can now use this transport plan to transfer the \ac{svm}-derived source scores to the individual target samples:

\begin{equation}
s_j^\T = \frac{1}{N}\sum_{i=1}^{n_\S}s_i^{\S}\delta(\gamma_{ij}>0)
\label{eq:rankTransport}
\end{equation}

Here, $s_i^{\S}$ denotes the distance to the \ac{svm} hyperplane for the $i$th source sample and $\delta(\cdot)$ is the Kronecker delta, returning value 1 if the condition inside the brackets is true, and 0 otherwise. $s_j^{\T}$ is the score for the $j$th target sample. In essence, we assign a score to each target sample as the sum of the \ac{svm} hyperplane distances of those \emph{source} samples whose \ac{ot} link ($\gamma_{ij}$) is non-zero, normalized by $N = \sum_{i=1}^{n_\S}\delta(\gamma_{ij}>0)$. An exemplar mapping on a subset of the training data is shown in Figure~\ref{fig:OTmapping}. This figure shows samples predicted by the \ac{cnn} that has been trained on the source (left point cloud), but not yet adapted to the target domain (right point cloud). The gray lines show links obtained by \ac{ot}\footnote{For illustration purposes we only show links that point to true positives in the target domain}. At a first glance, it is evident that the \ac{cnn} predicts orders of magnitude more false positives in the target domain, which is due to the domain shift between the two datasets. If we follow the \ac{ot} links from all \emph{source} true positives, we hit 51 target true positives (around 10\% of the target true positives) and 759 target false positives (around 2.6\% of the target false positives). This may sound like a low-precision result, but note that we prioritize the source true positives with our \ac{ts} metric, thus drastically reducing the number of false positives (see results below). Also, the \ac{ot} links to the true positive target samples consistently come from true positive source samples, which indicates that the \ac{ot}-derived transport plan is globally sound and succeeds in mapping correct predictions together. In the end this means that \ac{ts} is particularly robust to class imbalances: even if the ratios of true to false positives differ substantially between the two domains (as is the case in our study), \ac{ts} still prioritizes the most confident predictions. Once the costs are transported to the target domain, we only need to rank the target samples and can further sample them with priority on high-quality predictions.

\subsection{Window cropping for patch-based labeling} \label{sec:windowCropping}

The second major component of our model, the window cropping strategy, extends the queried candidate with its spatial surroundings. In other words, for all candidates identified through \ac{ts}, we crop a patch of fixed size around them and have the entire area labeled by the oracle, instead of the single prediction.

As mentioned above, we seek to find a trade-off between simply locating animals and \ac{cnn} updates. Window cropping enhances both objectives, as it increases the total amount of labels obtainable from the oracle in a single query. We can crop a window around a query position in a \ac{uav} image in such a way that it includes as many other predicted candidates as possible. In the case of false positives, this increases the information flow to the \ac{cnn} and results in it making less false predictions during the next \ac{al} iteration. However, in case neighboring candidates are also true positives, window cropping can accelerate the retrieval rate of animals with minimal additional effort from the oracle. This is not unlikely, since animals tend to flock together in groups. If the \ac{cnn} thus finds just one of the animals in a herd, it is trivial for humans to localize the rest close-by this way.

\begin{figure}[ht]
\centering
\begin{subfigure}{0.39\linewidth} \centering
\includegraphics[width=0.9\textwidth]{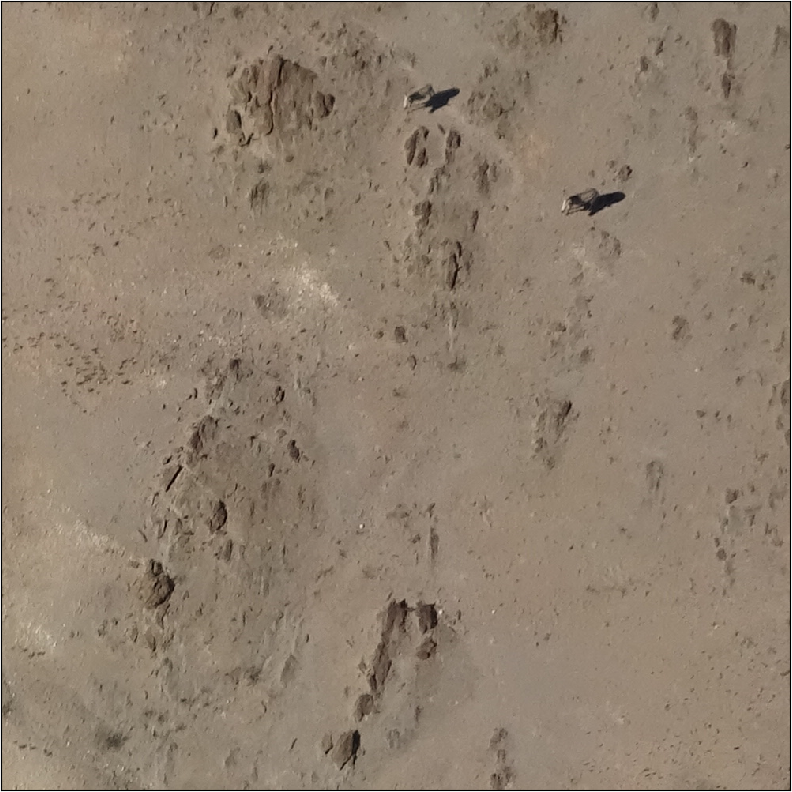}
\caption{}
\label{fig:neighborSampling}
\end{subfigure}
\begin{subfigure}{0.39\linewidth} \centering
\includegraphics[width=0.9\textwidth]{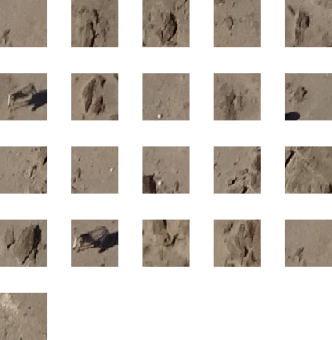}
\caption{}
\label{fig:neighborSampling_noCandidates}
\end{subfigure}
\caption{Patch of a target image (left) and all candidates predicted by the source \ac{cnn} in it (right). By cropping a larger patch around multiple candidates at once (left), the labeling process is both faster and more intuitive for human operators than querying on a per-candidate basis (right).}
\end{figure}

A further advantage of including the neighborhood lies in the ability of humans to be able to instantly recognize targets, if spatial context is provided. Consider Figures~\ref{fig:neighborSampling} (sample target image) and~\ref{fig:neighborSampling_noCandidates} (predicted candidates in it): querying the oracle for every candidate individually would not only be too exhaustive, but also more difficult, as the recognizability of the target depends heavily on spatial context, which might be missing (or confusing) on a per-sample query. In turn, locating animals in an adequately sized patch is a much simpler task for humans.

\begin{figure*}[ht]
\centering
\begin{subfigure}{0.3\linewidth} \centering
\includegraphics[width=\textwidth]{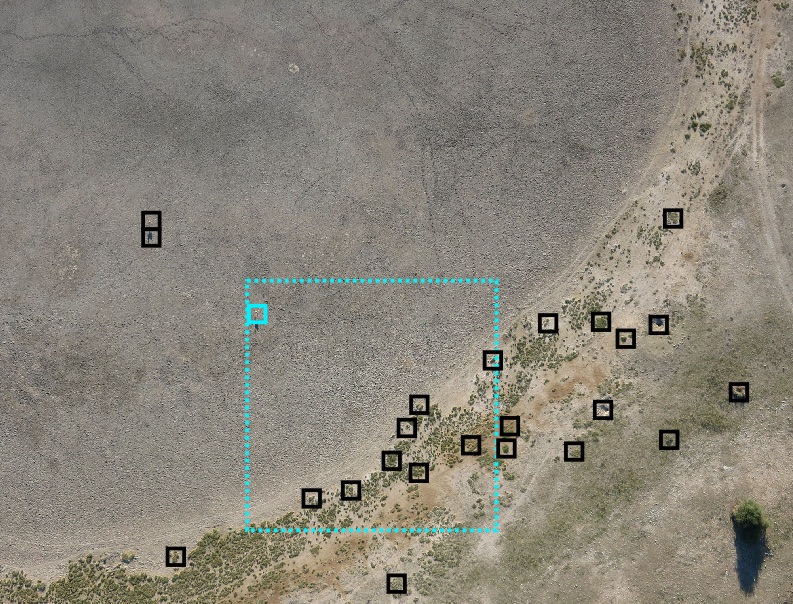}
\end{subfigure}
\begin{subfigure}{0.3\linewidth} \centering
\includegraphics[width=\textwidth]{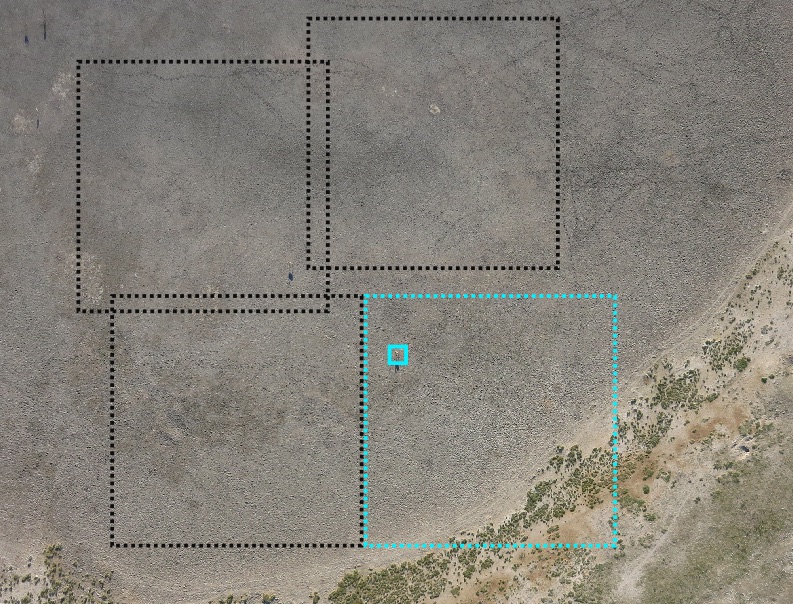}
\end{subfigure}
\begin{subfigure}{0.3\linewidth} \centering
\includegraphics[width=\textwidth]{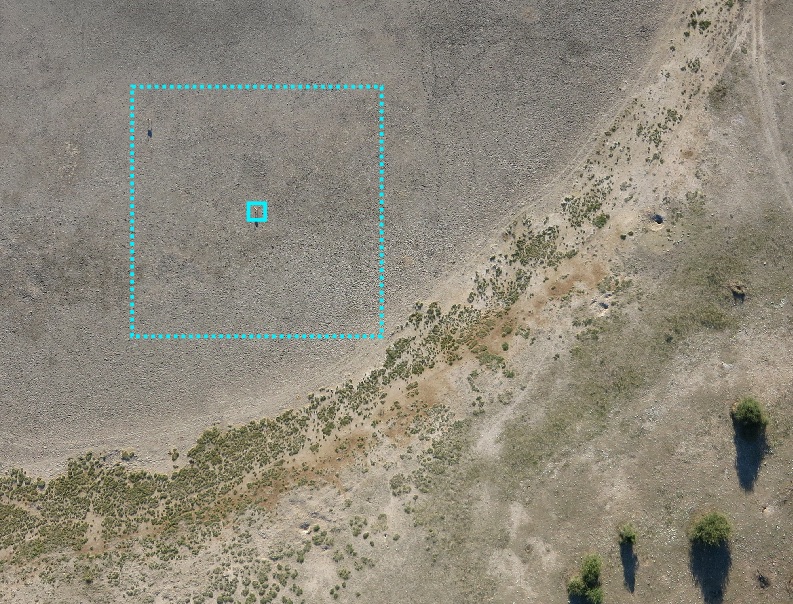}
\end{subfigure}
\caption{For window cropping, we address different scenarios to maximize the query gain: in the first situation (left), we place the candidate window (cyan dotted) so that it includes the anchor point selected by the querying strategy (cyan solid) and as many other candidates (black) as possible. In the second situation (middle), we minimize the overlap with previously queried windows (black dotted). If neither other candidates nor previous windows are present (right), we position the window centered around the anchor point.}
\label{fig:windowCropping}
\end{figure*}

It thus makes sense to crop patches in such a way that they include as many neighboring candidates as possible. We first define a patch rectangle as $r = \{r_x,r_y,r_w,r_h\}$, with $r_x$ and $r_y$ denoting the top-left corner of it, and $r_w$ and $r_h$ the width and height in pixels, respectively. Also, let $l = \{l_x,l_y\}$ be the position in the image of the candidate selected by the \ac{al} criterion, further referred to as the \emph{anchor point}. To select the best rectangle around the anchor point, we optimize a function that maximizes the number of candidates in the patch, minimizes the overlap with previously cropped patches and keeps the current candidate as close as possible to the center of the patch window:

\begin{equation}
\begin{split}
r^* = \argmin\limits_{r \in \R_{l}} \big((1-N(p,r)) + \max(I(r,\R_{q})) + \lambda \norm{r_c - l}^{2} \big)
\end{split}
\label{eq:windowCropping}
\end{equation}

where $\R_{l}$ is the set of windows that contain the anchor point $l$. The first term, $N(p,r)$, is the number of candidates $p$ inside rectangle $r$, normalized by the total number of candidates present in the image. The second term, $\max(I(r,\R_{q}))$, denotes the maximum area intersection between rectangle $r$ and all the rectangles in this image that have been queried before ($\R_{q}$). This term is normalized by the area of the rectangle so that it also sums to one, like the first term. The third term compares the anchor point $l$ with $r_c = \{r_x + r_w / 2, r_y + r_h / 2\}$, i.e. the center of the rectangle, by means of a norm and favors centering the window on the anchor. This last term primarily plays a role when the image only contains the anchor point $l$ (i.e., there are no other candidates nor any previously queried rectangles); hence, it is downweighted with a constant $\lambda$ (set to 0.01 in the experiments). Example scenarios for the three terms are shown in Figure~\ref{fig:windowCropping}. This score function is non-differentiable in multiple ways. However, since we restrict $r$ to always contain anchor point $l$, the search space $\R_{l}$ is very limited. We thus employ an exhaustive grid search around the anchor point.

We then query the oracle with this patch and receive positions of animals within, if present. Any other location in the image is labeled as background.

This procedure naturally depends on the patch size, where a compromise must be found: too large patches make it increasingly harder for humans to label, while too small patches exhaust the querying budget too quickly and provide less context. In this study, we limit experiments to a simulated oracle, but nevertheless use a patch size that we found reasonable while manually labeling the dataset. In detail, we crop patches of $1000 \times 1000$ pixels (approx. $60 \times 60$m) that provide a sufficiently large number of samples while still being easy for humans to label.

\section{Experiments} \label{sec:experiments}
We now put the proposed workflow to the test and describe the data and parameters below. Section~\ref{sec:dataset} describes the two datasets used; Section~\ref{sec:modelSetup} highlights parameters of the detector \ac{cnn} and of the \ac{al} routine.

\subsection{Study Area and Data} \label{sec:dataset}
We evaluate our proposed method on \ac{uav} datasets acquired over the Kuzikus wildlife reserve in Namibia\footnote{\url{http://kuzikus-namibia.de/xe_index.html}}. Kuzikus is a private-owned park in the African savanna and home to multiple species of large mammals like kudus, giraffes, zebras, black rhinos, and more. In total, more than 3000 individuals are spread across an area of $\mathrm{103km^2}$~\cite{Rey2017,Kellenberger18}.

In 2014 and 2015, two image acquisition campaigns were carried out by the SAVMAP consortium\footnote{\url{https://lasig.epfl.ch/savmap}}. A SenseFly eBee\footnote{\url{https://www.sensefly.com}}, equipped with a consumer-grade RGB compact digital camera, was employed for both campaigns. This resulted in 654 images for the 2014 campaign, and 3254 for the year 2015. The images of the first acquisition were initially labeled in a crowd-sourcing operation organized by MicroMappers\footnote{\url{https://micromappers.wordpress.com}}~\cite{ofli2016combining}, followed by several iterations of refinement by the authors. The 2015 images were completely labeled by the authors.

The final statistics for both datasets are listed in Table~\ref{tab:kuzikusData}. Although both datasets were acquired over geographically overlapping areas, they feature a substantial domain shift in multiple ways: in terms of \emph{external conditions}, the datasets were acquired at different times of the year (May 2014, resp. February and May 2015), under different weather and lighting conditions, with different cameras and varying flying altitudes above ground.
Furthermore, additional shifts can be observed in the \emph{label space}: the 2014 data already have a substantial class imbalance (1:10'000 in terms of animal-to-background pixels), but the 2015 dataset is larger and even more imbalanced, with an overall lower proportion of animals.

We divided the source (2014) dataset into training, validation and test splits according to the following set of rules:
\begin{itemize}
\item We assign entire images to only one of the three sets to avoid autocorrelation effects.
\item We differentiate between images that contain at least one animal and empty ones. All the images with at least one animal are distributed so that the number of \emph{animals} in the sets are distributed as follows: 70\% for the training set, 10\% for validation, and 20\% for testing.
\item All the remaining images (i.e., those without any animal) are then distributed at random to meet the same 70-10-20 split, but this time based on the number of \emph{images}, as closely as possible.
\end{itemize}

For the target (2015) data, we do not require a test set, since we sample directly using the oracle. However, we do use a small validation set for hyperparameter fine-tuning. Details on the dataset splits can be found in Table~\ref{tab:kuzikusDataSplits}.

\begin{table}
\centering
\caption{Overview of the 2014 and 2015 Kuzikus \ac{uav} datasets.}
\label{tab:kuzikusData}
\begin{tabular}{c | c c}
\hline
& Set 1 & Set 2\\
\hline
Year & 2014 & 2015\\
Image sizes & $4000\times3000$ & $4896\times3672$,\\
& & $4608\times3456$\\
Camera models & Canon PowerShot S110 & Sony DSC-WX220,\\
& & Canon IXUS 127 HS\\
No. images & 654 & 3254\\
with animals & 239 & 111\\
without & 415 & 3143\\
No. animals & 1183 & 646\\
Elevation a.g. (est.) & 120m & 160m\\
\hline
\end{tabular}
\end{table}

\begin{table*}
\centering
\caption{Split properties for the 2014 and 2015 datasets.}
\label{tab:kuzikusDataSplits}
\begin{tabular}{r || c c c | c | c c c | c | c c c | c}
\hline
Set & \multicolumn{4}{c|}{training} & \multicolumn{4}{c|}{validation} & \multicolumn{4}{c}{test}\\
& \multicolumn{3}{c|}{No. images} & No. animals & \multicolumn{3}{c|}{No. images} & No. animals & \multicolumn{3}{c|}{No. images} & No. animals\\
& with animals & without & total & & with animals & without & total & & with animals & without & total &\\
\hline
2014 & 159 & 291 & 450 & 830 & 35 & 41 & 76 & 118 & 45 & 83 & 128 & 235\\
2015 & 91 & 2750 & 2841 & 565 & 20 & 393 & 413 & 81 & - & - & - & -\\
\hline
\end{tabular}
\end{table*}

\subsection{Model Setup} \label{sec:modelSetup}

In the following, we highlight the main model components and their parameters: Section~\ref{sec:cnnTraining} explains the deep \ac{cnn} used for animal detection, and Section~\ref{sec:alLoop} provides details on the \ac{al} framework.

\subsubsection{\ac{cnn} Training} \label{sec:cnnTraining}

In this study, we follow the training recommendations presented in~\cite{Kellenberger18}, which are specifically tailored to animal censuses in heavily imbalanced datasets. These recommendations include class weights; a special ``border'' class that is placed in the 8-neighborhood of an animal location to reduce multiple detections of it; curriculum learning, where the model is first trained on images that always contain an animal; and hard negative mining, which amplifies the weights of the four most confidently predicted false alarms after epoch 80.

\begin{figure}
\includegraphics[width=\columnwidth]{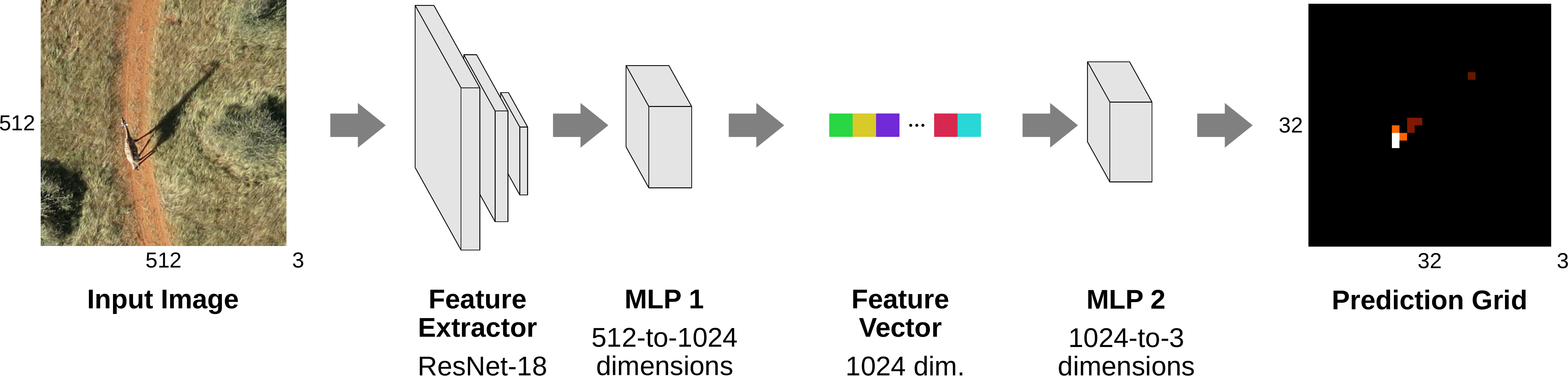}
\caption{Basic architecture for our animal detector, following~\cite{Kellenberger18}. We employ the main blocks of a ResNet-18 pretrained on ImageNet and add two more \acp{mlp} and ReLU nonlinearities. The feature vector after the \ac{mlp} 1 is used by \ac{ot} within \ac{ts}.}
\label{fig:CNN_architecture}
\end{figure}

We further adopt the detector \ac{cnn} from~\cite{Kellenberger18}, whose architecture is shown in Figure~\ref{fig:CNN_architecture}. As a feature extractor, it employs a ResNet-18~\cite{He2015} that had been pre-trained on the ILSVRC dataset~\cite{Russakovsky2015}. The model accepts image patches of size $512\times512$ and predicts a downsampled grid of animal probabilities ($32\times32$). To adapt it for predicting a grid of this size instead of a single label, the last layers, including the global average pooling layer at the end, are removed and the first convolutional layer's stride reduced to 1. We use the 512-dimensional feature vectors from the basic model (i.e., the output of the last residual block) for further adaptations. For obtaining per-location class predictions, two \acp{mlp} map the feature vectors from 512 to 1024 dimensions, and then to the 3 classes (background, animal, border), respectively. Furthermore, to avoid dependencies on mini-batch configurations both during training and testing, we replace all batch normalization layers with instance normalization~\cite{Ulyanov2017}, which performs unit-norm scaling for each image in the respective mini-batch individually. We found this substitution not to harm the model performance, but to help stabilize prediction consistency.

For the fine-tuning stages throughout the \ac{al} iterations, we lower the learning rate from $\mathrm{10^{-6}}$ (used on source) to $\mathrm{10^{-7}}$---we found this to prevent oscillation effects on the reduced-sized target training set. Also, we disable curriculum learning and train on the full growing \ac{al}-derived dataset at every iteration. All other parameters and training procedures, such as hard negative mining, are kept the same as in the source model.

\subsubsection{\ac{al} Loop} \label{sec:alLoop}

In a pre-stage, we train our detector \ac{cnn} until convergence on source. We use this \ac{cnn} to obtain all candidates in the source training set, keeping all predictions with animal confidence of 0.1 or higher. To reduce the number of double-predictions, we employ \ac{nms} with a search radius of 2 prediction grid cells, retaining only candidates in a 4-neighborhood with the highest confidence.

Next, we run a total of ten \ac{al} loops, querying 50 patches of $\mathrm{1000 \times 1000}$ pixels size per iteration. We believe 50 patches to be a fair amount to query without risking the human annotators to make errors due to fatigue (note that other studies used query sizes of up to 200 images in more heterogeneous datasets~\cite{Kao2018}). Using those we fine-tune the \ac{cnn} on the target training set for 12 epochs per \ac{al} iteration, which we deem a reasonable trade-off between training time and accuracy gain. Since we queried patches that are larger than what the \ac{cnn} accepts as an input, we can perform extra data augmentation by randomly cropping sub-patches of the labeled areas.

We compare our \ac{ts} strategy to three baselines: random candidate selection, Breaking Ties~\cite{luo2005active}, and \ac{cnn} confidence for being an animal (``max confidence''), all with window cropping enabled. Since model fine-tuning and candidate re-prediction is computationally intensive, we assess two scenarios for each sampling strategy: one with \ac{cnn} fine-tuning at every \ac{al} iteration, and one without (i.e., using only the source model to predict candidates once and querying with \ac{ts} only on those samples).

\section{Results and Discussion} \label{sec:results}

\begin{figure}
\centering
\includegraphics[width=0.65\linewidth]{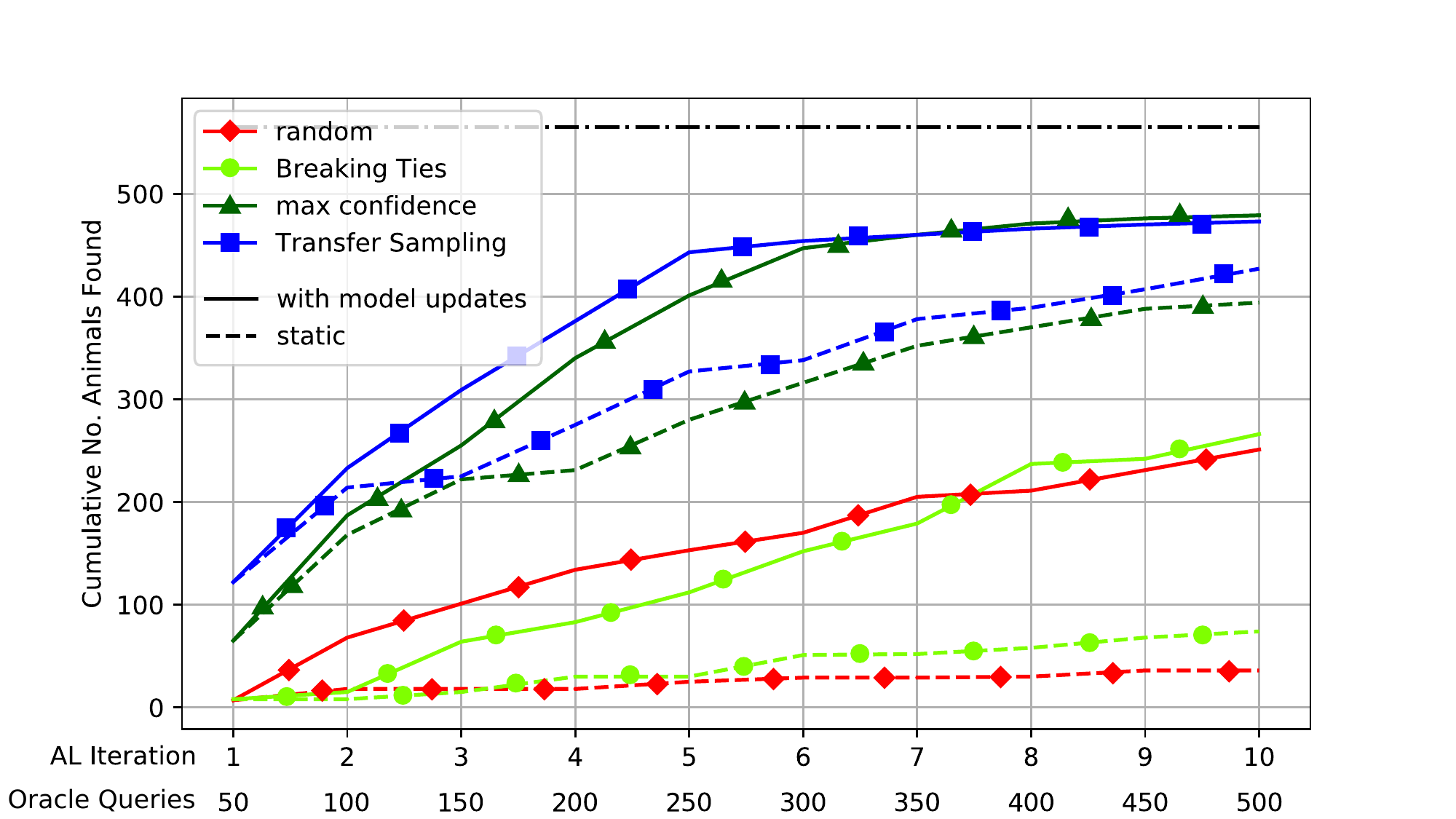}
\caption{Cumulative number of animals found over the \ac{al} iterations. Solid lines denote the criteria performances with model updates and target candidate re-predictions at every iteration; dashed lines are the static performances (continuous sampling on the initially predicted candidates). The black dash-dotted line marks the total number of animals in the target training set. Best viewed in color.}
\label{fig:finalResults_numAnimalsFound}
\end{figure}

\begin{figure*}
\centering
\begin{subfigure}{5mm} \centering
{\rotatebox[origin=t]{90}{\large{\ac{ts}}}}
\end{subfigure}
\begin{subfigure}{0.45\linewidth} \centering
\includegraphics[width=\textwidth]{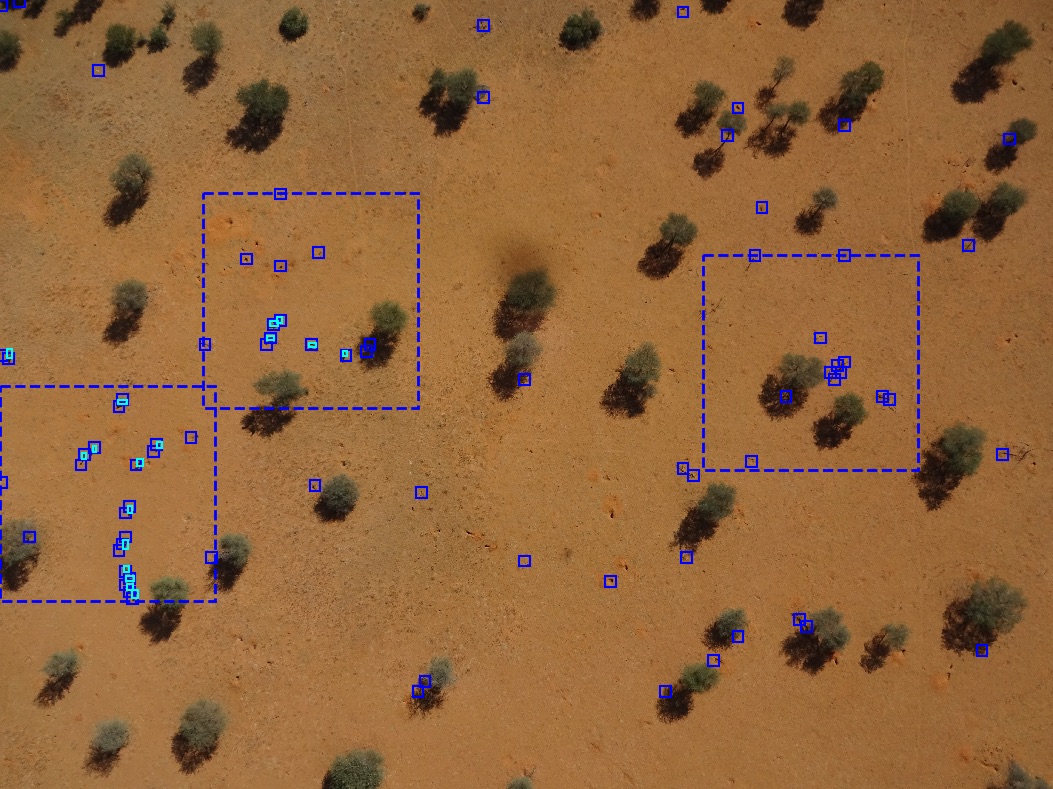}
\end{subfigure}
\hspace{5mm}
\begin{subfigure}{0.4\linewidth} \centering
\includegraphics[width=\textwidth]{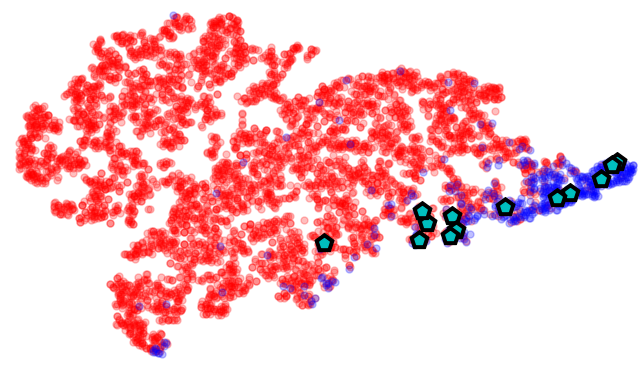}
\end{subfigure}
\\
\vspace{2mm}
\begin{subfigure}{5mm} \centering
{\rotatebox[origin=t]{90}{\large{max confidence}}}
\end{subfigure}
\begin{subfigure}{0.45\linewidth} \centering
\includegraphics[width=\textwidth]{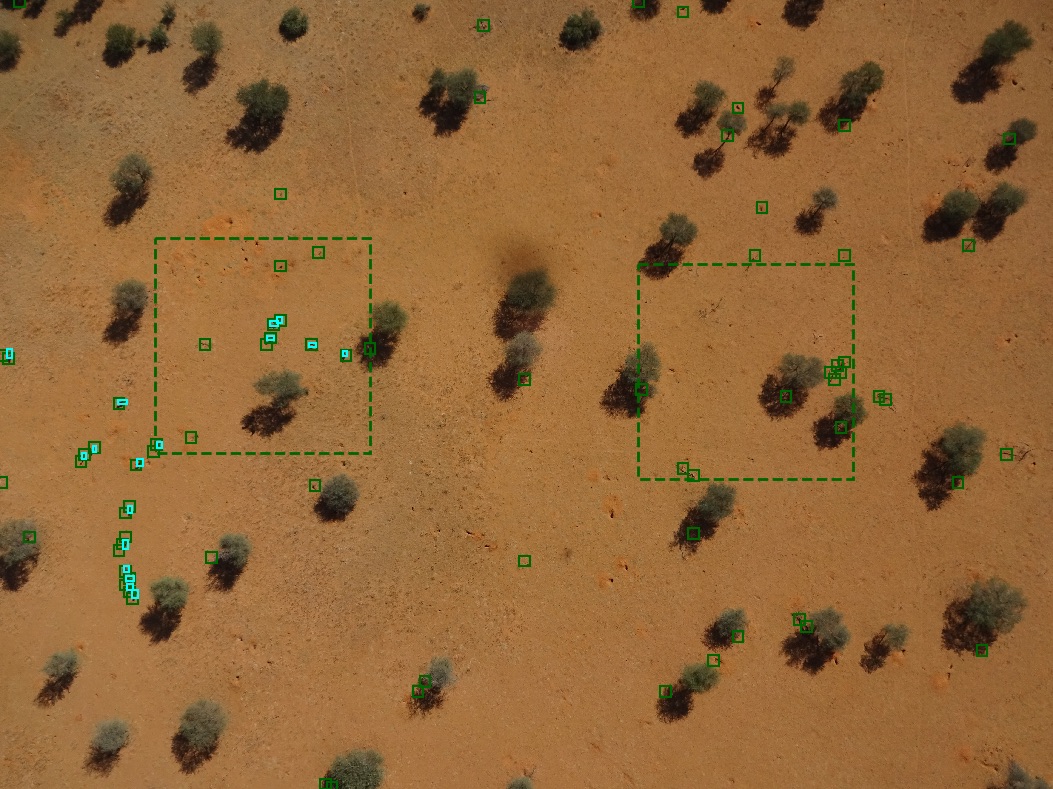}
\end{subfigure}
\hspace{5mm}
\begin{subfigure}{0.4\linewidth} \centering
\includegraphics[width=\textwidth]{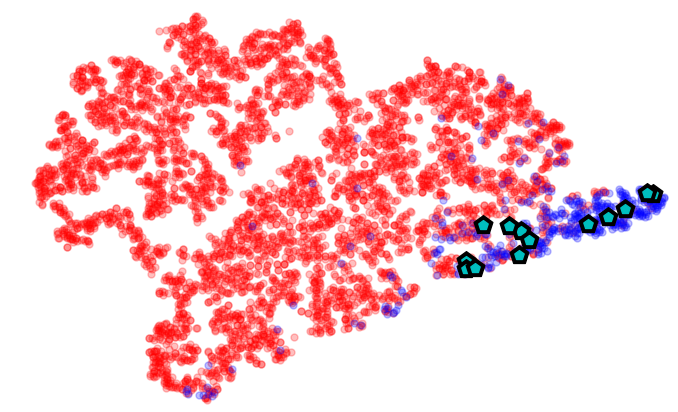}
\end{subfigure}
\caption{Prediction examples for the \ac{ts} (top) and max confidence (bottom) strategies after five \ac{al} iterations (250 queries), together with their respective t-SNE plots of the predicted candidates (right). Cyan pentagons in the t-SNE plot correspond to the cyan ground truth bounding boxes in the \ac{uav} images.}
\label{fig:visualExamples_1}
\end{figure*}

Figure~\ref{fig:finalResults_numAnimalsFound} shows the number of animals found over the course of the ten \ac{al} iterations. Already after the first 50 queries, \ac{ts} found 122 animals and is significantly ahead of the baselines. This trend continues throughout the iterations, and after five \ac{al} iterations, \ac{ts} found 443 out of the total 565 animals (78.4\%). At this stage, the oracle had been queried 250 times. Afterwards, the total number of correctly identified animals slightly rises to 473 (83.72\%). Our window cropping algorithm allows sampling patches in an out-of-grid fashion. However, if we assume uniform sampling on a grid per image, the target training set would consist of 54'324 queryable patches. This means that \ac{ts} only requires the user to review around half a percent of the dataset in order to find almost 80\% of the animals.

In comparison to all baselines, \ac{ts} manages to yield a higher recall almost throughout the entire process. Although the max confidence ranking manages to reach roughly the same level, it does so only after the sixth \ac{al} iteration. \ac{ts} in turn identified the same number of animals already an entire iteration earlier, and stayed above the rest until then by quite a margin. This means that substantially less queries need to be made to the annotator when using \ac{ts}, resulting faster convergence and hence a more economical retrieval process.

Figure~\ref{fig:visualExamples_1} shows the selected patches, predicted candidates and ground truth for both \ac{ts} (top image) and max confidence (bottom): the latter does manage to find a reasonable number of true positives, but nonetheless misses more than half of the animals present in the scene. Explanation may be found in the t-SNE plots (right side of each image), which indicate that most of the true positives are to be found primarily in one area in the bottom right of the feature space. One might expect a correlation between feature space locations and confidences, but as shown here, this is only partially the case: \ac{ts} manages to get ahead of max confidence by strictly sampling in this area of high true positives concentration (bottom right of t-SNE plots), instead of according to confidences.

\begin{figure*}[ht]
\centering
\begin{subfigure}{0.9\textwidth} \centering
\includegraphics[width=0.8\textwidth]{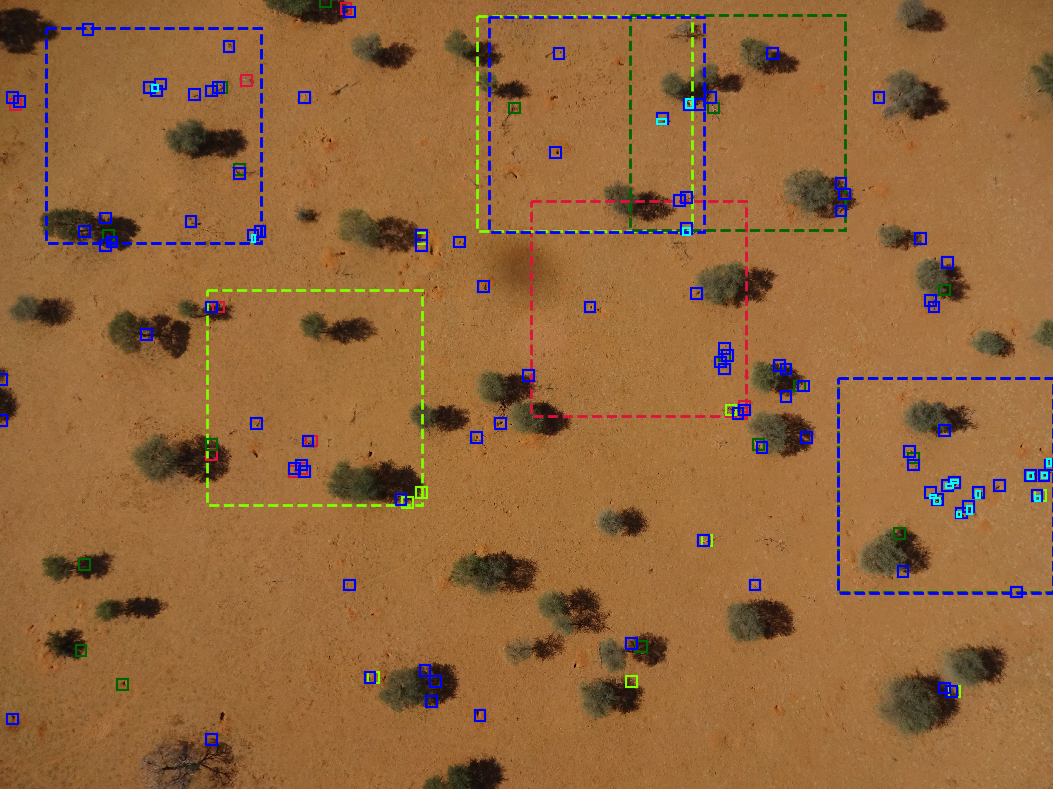}
\end{subfigure}\\
\vspace{2mm}
\begin{subfigure}{0.9\textwidth} \centering
\includegraphics[width=0.8\textwidth]{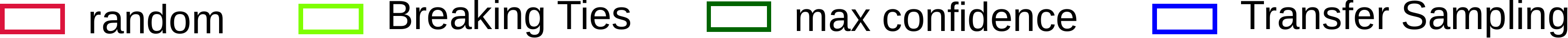}
\end{subfigure}
\caption{Example image from the target training set with annotations after five \ac{al} iterations, showing all selected patches per sampling strategy (dashed), the candidates predicted by each \ac{cnn} at the very last \ac{al} iteration, as well as the ground truth (cyan rectangles). Since only the last predictions, but all patch rectangles are shown, some of them (i.e., from earlier \ac{al} iterations) do not encompass any candidate.}
\label{fig:visualExamples_combined_1}
\end{figure*}

The performance of the other baselines is less satisfactory. For instance, the focus of Breaking Ties lies on minimizing the model's uncertainty. It is therefore unsurprising that it fails even worse than random sampling in finding animals. However it is still able to locate some of the animals. Consider Figure~\ref{fig:visualExamples_combined_1}, which shows selected patches and predicted candidates on an example for all strategies. In this case, random sampling managed to find one of the animals in an earlier iteration, but misses the rest. Breaking Ties sampled two times in the image and found three animals, but this was most likely due to neighboring candidates that do not represent true positives (note the candidates predicted by Breaking Ties lying all on transition areas from shadow to ground). Max confidence again found the same three animals, but missed the hotspot in the center-right portion of the figure. Lastly, only \ac{ts} managed to locate every single animal present, and did so with a minimal number of queries. In essence, the other strategies occasionally show true positives, but mainly due to window cropping.

Finally, in all cases we observe a significant boost when the model is updated and candidates are re-predicted versus the static ``one-shot'' prediction and sampling (Figure~\ref{fig:finalResults_numAnimalsFound}). This indicates that the labels provided by the oracle at each stage are useful and complete enough for adapting the \ac{cnn} to the target domain, so that the predicted candidates at the next iteration are of higher quality. Window cropping helps particularly in this case, since \acp{cnn} require a large number of training samples: on the one hand, it increases the amount of image data seen by the model. On the other, it also increases the information gain: since the cropping strategy maximizes the number of predicted candidates to be included per patch, it maximizes both the number of true and false positives in the new dataset. In this respect both true and false positives are vital for the model, improving its abilities to better separate animals from background in the target dataset.

Noteworthy in this context is the difference in the number of predicted candidates between \ac{ts} and the other strategies. After the fifth \ac{al} iteration, the \ac{cnn} trained with \ac{ts} produced a significantly higher number of candidates than all the baselines, which can be seen by the high number of blue squares in Figure~\ref{fig:visualExamples_1}. A possible explanation for this phenomenon is that \ac{ts} finds more animals already in the first \ac{al} iteration: the \ac{cnn} is then fine-tuned with a lot more true positives and therefore predicts more candidates when re-applied to the target dataset. In a fully-automated evaluation setting (i.e., without any oracle input), this could be problematic, since this increased number of predictions also results in more false positives. However, thanks to \ac{ts} we can filter the predictions very efficiently, and as a result, localize the animals even in increasingly imbalanced settings.

\section{Conclusion} \label{sec:conclusion}
In this paper, we have studied the task of repeated animal censuses on \ac{uav} imagery by addressing the domain adaptation problem involved. To do so, we integrated a deep \ac{cnn}-based animal detector in an \ac{al} loop. The core component of our strategy is the \ac{al} criterion: unlike traditional approaches that seek to maximize the model performance on the new dataset, our \acreset{ts}\ac{ts} criterion is designed to localize the rare animals in tens of thousands of false alarms as efficiently as possible. \ac{ts} works by leveraging the superior performance of the \ac{cnn} detector in the source dataset (which it had been trained on) and transferring this knowledge to the target set using the distribution-mapping framework \acreset{ot}\ac{ot}. The number of hits was further raised by integrating a smart window cropping strategy that maximizes the number of detections to be labeled per query, while making the labeling process itself more intuitive. Our experiments in the Namibian natural reserve Kuzikus have shown that \ac{ts} indeed outperforms other \ac{al} criteria by a large margin and allows retrieving 78.4\% of the animals in just 250 queries, resp. by having the oracle review less than half a percent of the entire dataset. In effect, this method thus allows for efficient and economic repetitions of animal censuses as it integrates both the adaptation and required manual verification stages into one optimized, interactive workflow.

Future work may extend this concept in multiple ways: for example, experiments with human annotators instead of simulated oracles would highlight requirements by park rangers. Extending \ac{ts} e.g. by a measure of the user's confidence in providing a ground truth~\cite{tuia2013learning} could improve the real-world applicability of such a system. On a different track, adaptations to other geographical areas instead of new acquisitions over the same game reserve would allow testing the strategy under potentially even stronger domain shifts.


\section*{Acknowledgments}
This  work  has  been  supported  by  the  Swiss  National  Science  Foundation  grant  PZ00P2-136827  (DT,
\url{http://p3.snf.ch/project-136827}). The authors would like to acknowledge the SAVMAP consortium (in particular
Dr. Friedrich Reinhard of Kuzikus Wildlife Reserve, Namibia) and the QCRI and Micromappers (in particular Dr. Ferda Ofli and Ji Kim Lucas) for the support in the collection of ground truth data. We gratefully acknowledge the support of the NVIDIA Corporation with the donation of the Titan V GPU used for this research.

\bibliographystyle{IEEEtran}
\bibliography{main}

\end{document}